\documentclass[11pt]{article}

\usepackage{acl}
\usepackage{paralist}
\usepackage{amssymb}
\usepackage{graphicx}
\usepackage{xcolor}
\usepackage{tikz}
\usetikzlibrary{calc,positioning}
\usepackage{times}
\usepackage{latexsym}

\usepackage[T1]{fontenc}

\usepackage[utf8]{inputenc}
\usepackage{tikz}
\usepackage{pgfplots}
\usetikzlibrary{patterns}
\pgfplotsset{compat=1.18}
\usepackage{microtype}

\usepackage{inconsolata}

\usepackage{graphicx}

\usepackage{color}
\usepackage{array}
\usepackage{enumitem}
\usepackage{caption}
\usepackage{algorithm}
\usepackage{algorithmic}
\usepackage{amsmath}
\usepackage{amssymb}
\usepackage{graphicx}
\usepackage{multirow}
\usepackage{booktabs}
\usepackage{bm}
\usepackage{tikz}
\usepackage{pgfplots}
\usepackage{xspace}
\usepackage{paralist}
\usepackage{tabularx}
\usepackage{graphicx}
\usepackage{subcaption}
\usepackage{enumitem}
\usepackage[utf8]{inputenc}

\usepackage{microtype}

\usepackage{inconsolata}

\usepackage{color,xcolor}
\usepackage{colortbl}
\usepackage{arydshln}
\usepackage{soul}

\usepackage{tcolorbox}

\usepackage{adjustbox}
\tcbuselibrary{breakable}

\newcommand{\mcolorbox}[2]{%
  \begingroup\setlength{\fboxsep}{1pt}%
  \colorbox{#1}{\text{\hspace*{2pt}\vphantom{Ay}#2\hspace*{2pt}}}%
  \endgroup
}

  {\list{}{\leftmargin=0.3in\rightmargin=0.3in}\item[]}%
  {\endlist}

\definecolor{nmgray}{RGB}{229,229,229}
\definecolor{underlinegray}{RGB}{197,197,197}

\definecolor{introblue}{RGB}{0,176,240}
\definecolor{introgreen}{RGB}{0,203,134}
\definecolor{introgreen2}{RGB}{139,243,206}

\definecolor{lightyellow}{RGB}{255,251,239}

\newcommand{\cmark}{{\color{introgreen}\checkmark}}
\newcommand{\xmark}{{\color{black!35}\ensuremath{\times}}}

\newtcolorbox{promptbox}[1]{breakable, colback=gray!4, colframe=black!55,
  boxrule=0.5pt, arc=2pt, left=5pt, right=5pt, top=3pt, bottom=3pt,
  fonttitle=\bfseries\footnotesize, coltitle=white, colbacktitle=black!60,
  title={#1}, fontupper=\footnotesize\ttfamily, before upper={\obeylines\setlength{\parindent}{0pt}}}

\setul{0.3ex}{0.2ex}
\setulcolor{underlinegray}

\newtcolorbox{mybox}[2][]{
width=\columnwidth,
colback = nmgray!75!white, 
colframe = nmgray!75!white, 
boxsep=0pt,left=10pt,right=10pt,top=0pt,bottom=0pt,
fontupper=\linespread{0.9}\selectfont,
title=#2,#1}

\newtcolorbox{mnewbox}[2][]{
  enhanced,
  overlay unbroken and first={
    \node[anchor=north east,outer sep=0pt] at (frame.north east) {#2};
  },
  #1
}
\usepackage{paralist}

\author{
Jie Yang\textsuperscript{\rm 1},  \quad
Wenhao Xu\textsuperscript{\rm 1},  \quad
Shuhui Lin\textsuperscript{\rm 2},  \quad
Hao Fei\textsuperscript{\rm 3}\thanks{Corresponding author.}\\
\textsuperscript{\rm 1} National University of Singapore \quad \textsuperscript{\rm 2}  Tsinghua University \quad 
\textsuperscript{\rm 3}   University of Oxford
 \\
\tt{e1554543@u.nus.edu, haofei7419@gmail.com} 
}

\title{\textbf{\texttt{EmpaAva}}: An Open-source Agentic 3D-Avatar Empathetic Live Chatbot}

\begin{document}
\maketitle
\begin{abstract}
This paper presents \textbf{\texttt{EmpaAva}}, to our knowledge the first open-source, agentic 3D-avatar empathetic chatbot, which carries empathetic response generation (ERG) from text-only exchanges into live, face-to-face interaction. Through a video-call-like interface, a user speaks to a 3D digital human that reads their affect from speech and optional vision, and replies with emotional speech, lip-synced facial motion, and photorealistic 3D Gaussian rendering. At its core, an LLM coordinates a \textbf{Tri-Agent Architecture}, in which perception, empathetic response planning, and embodied rendering form a closed loop, paired with a \textbf{Response Planning} layer that compiles each reply into an executable multimodal plan, keeping voice, expression, and rendering on one empathetic intent. Building on strong open-source modules, \texttt{EmpaAva} supplies the intelligence that binds them into one controllable, inspectable experience. In automatic and human evaluations, EmpaAva surpasses text-only, 2D talking-face, and multimodal avatar baselines in emotion understanding, response quality, and audio-visual consistency.
We open-source EmpaAva\footnote{\url{https://empaava.top}} with an online live demo\footnote{\url{https://empaava.top/demo}}.
\end{abstract}

\vspace{-2mm}
\section{Introduction}
\vspace{-2mm}
The past few years have turned large language models~\citep[LLMs;][]{chatgpt,abs-2210-11416} into remarkably fluent conversational partners, yet fluency alone does not make a machine feel human. Human-level interaction also demands \emph{empathy}: the capacity to read another's feelings and reply with comfort and support. This goal drives empathetic response generation~\citep[ERG;][]{rashkin-et-al-2019-towards} and a line of research that infers a user's affect before answering~\citep{lin2019moel,majumder2020mime,li2020empdg}, already backing scenarios from mental-health companionship to emotional-support dialogue~\citep{esconv}.

Human empathy, however, travels through far more than words. A trembling voice, a softened gaze, a hesitant pause carry affective weight that plain text can never hold. Recent research therefore pushes ERG into a multimodal setting, jointly modeling what a user says, how they say it, and how their face changes~\citep{empathyear,avamerg}. In parallel, avatar-driven video generation has matured dramatically, with audio-driven portrait models~\citep{sadtalker,emo,vasa1} synthesizing lifelike talking faces. A face-to-face digital human is arguably the most natural vessel for empathy: it lets a user \emph{see} the listener nod, smile, and react, closing a loop that a chat bubble leaves open.

\begin{figure*}[!t]
\centering
\includegraphics[width=0.93\textwidth]{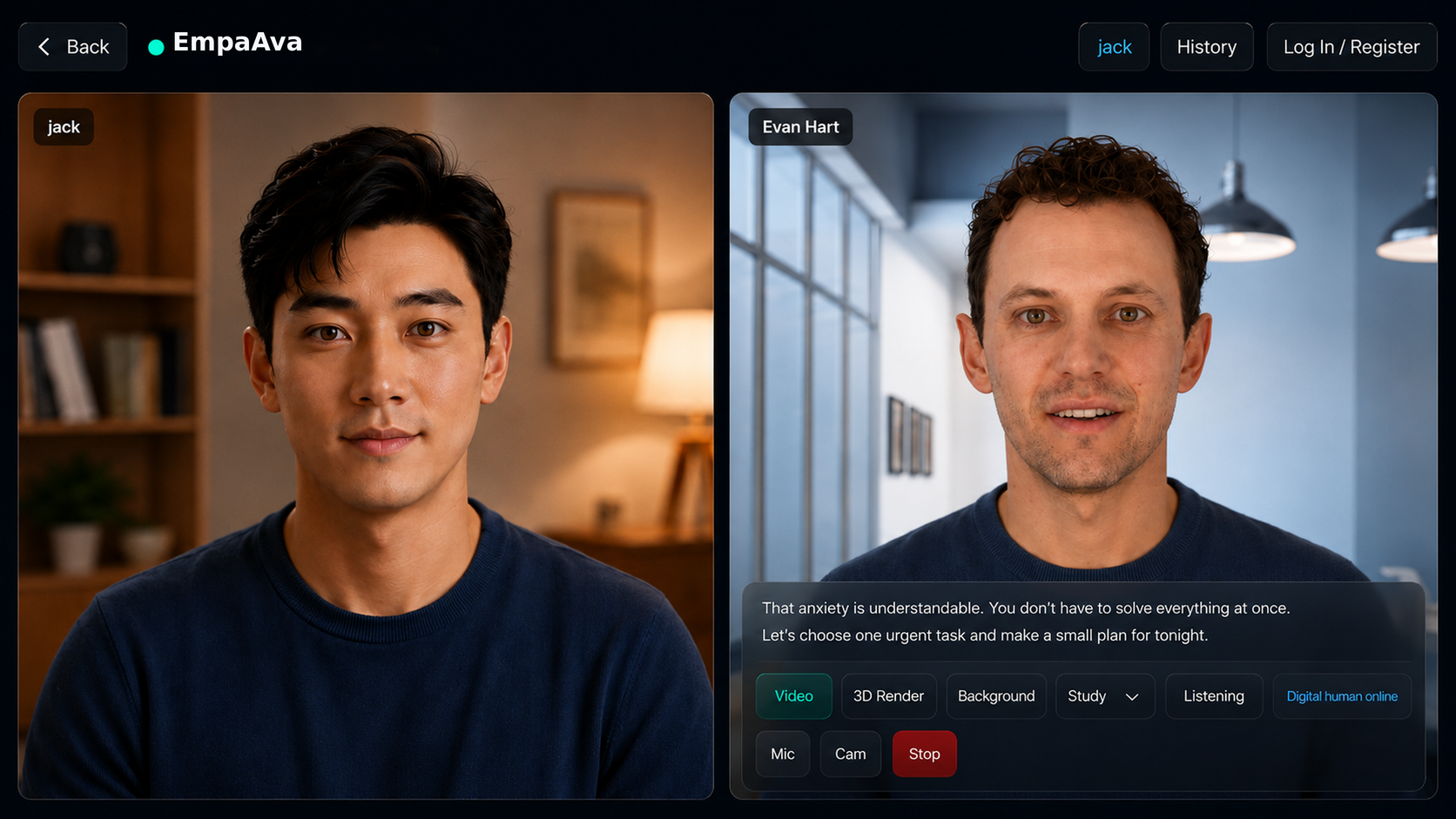}
\caption{EmpaAva. In a video-call-like booth, the user speaks to a 3D digital human that senses their emotion and replies face to face, with emotional speech, lip-synced facial motion, and photorealistic 3D-avatar rendering.}
\label{fig:teaser}
\vspace{-10pt}
\end{figure*}

Turning this vision into a deployable system, though, is far from settled, and two obstacles stand out.
The first is \textbf{expressive fidelity}. Many empathetic-avatar efforts still rest on 2D talking-face synthesis~\citep{wav2lip,sadtalker}, which copes poorly with three-dimensional consistency, natural head movement, and spatial realism. For a system meant to sit across from a user in a video call, a 3D avatar is the more faithful embodiment: FLAME~\citep{flame} offers controllable geometry and expression, while 3D Gaussian Splatting~\citep{3dgs,gaussianavatars,gaussiantalker} renders photorealistic, view-consistent heads beyond the reach of 2D pipelines.
The second is \textbf{interaction intelligence}. Existing 3D-avatar methods mostly animate a talking head from given speech, and stop short of the full arc from perceiving a user's emotion, to planning an empathetic reply, to voicing it through an embodied face. No \emph{open-source, live} system, to our knowledge, wraps this arc into something the community can readily run and extend.

We close this gap with \textbf{\texttt{EmpaAva}}, to our knowledge the first open-source, agentic 3D-avatar empathetic live chatbot. As previewed in Figure~\ref{fig:teaser}, a user steps into a video-call-like booth and talks with a 3D digital human that listens, gauges the user's affect, and answers with synchronized speech, facial motion, and emotionally consistent rendering. For perception and generation, EmpaAva stands on strong open-source shoulders, e.g., 3DGS+FLAME rendering and the emotional TTS engine EmotiVoice~\citep{emotivoice}, so that its contribution centers on the intelligence orchestrating these parts.

That intelligence is arranged as an agentic system with an LLM at its reasoning core, echoing the shift toward LLMs as controllers that plan and dispatch specialized tools~\citep{react,hugginggpt,llmagents}. Two designs carry the novelty. The \mcolorbox{introgreen2}{\textbf{Tri-Agent Architecture}} splits the long \textit{perceive}--\textit{decide}--\textit{express} chain into three cooperating agents, \textsc{PerceptionAgent}, \textsc{ResponseAgent}, and \textsc{RenderAgent}, mirroring how a person first reads a situation, then decides how to respond, and finally conveys it through voice and expression. Beyond keeping each stage modular and independently upgradable, the split turns a flat pipeline into a closed empathetic loop that tracks the user's emotional state across turns.

The second is a \mcolorbox{introgreen2}{\textbf{Response Planning}} layer between the LLM and the rendering back-end. Rather than forward a bare line of text, ResponseAgent emits a structured \emph{reply plan} fixing who speaks, in which voice and emotional tone, against which background, and which modules to invoke. This plan acts as one shared expressive intent that every generator obeys, keeping the comforting wording, the gentle voice, and the softened expression aligned, and granting fine control over how strongly the avatar smiles, nods, or slows its pace.

Overall, we make the following contributions:

\setdefaultleftmargin{1.5em}{1.0em}{1.87em}{1.7em}{1em}{1em}
\begin{compactenum}[1)]

\item We present \texttt{EmpaAva}, to our knowledge the first \textbf{open-source, live} agentic 3D-avatar empathetic chatbot, delivering face-to-face empathetic interaction over text, speech, and vision.

\item We propose a \textbf{Tri-Agent Architecture} that decomposes empathetic interaction into perception, response planning, and embodied rendering, forming a closed, controllable, and state-aware empathetic loop.

\item We introduce a \textbf{Response Planning} layer that translates an LLM reply into an executable multimodal expression plan, enforcing cross-modal consistency and controllable avatar behavior.

\item We release the full implementation, an online live demo, and the evaluation setup to support research on embodied empathetic agents.

\end{compactenum}

\begin{table}[!t]
\centering
\footnotesize
\setlength{\tabcolsep}{4.2pt}
\renewcommand{\arraystretch}{1.15}
\caption{Positioning of \texttt{EmpaAva} against representative prior work. \textbf{OS}: open-source; \textbf{LLM}: LLM-based reasoning core; \textbf{MM}: multimodal (text{+}speech{+}vision) input; \textbf{3D}: embodied 3D-avatar output; \textbf{Emp.}: empathy-oriented; \textbf{Live}: interactive live system.}
\label{tab:related}
\resizebox{\columnwidth}{!}{%
\begin{tabular}{lcccccc}
\toprule
\textbf{System} & \textbf{OS} & \textbf{LLM} & \textbf{MM} & \textbf{3D} & \textbf{Emp.} & \textbf{Live} \\
\midrule
Text ERG~{\scriptsize\citep{lin2019moel}}       & \cmark & \xmark & \xmark & \xmark & \cmark & \xmark \\
2D Talking-Face~{\scriptsize\citep{sadtalker}}  & \cmark & \xmark & \xmark & \xmark & \xmark & \xmark \\
3D Avatar~{\scriptsize\citep{gaussianavatars}}  & \cmark & \xmark & \xmark & \cmark & \xmark & \xmark \\
AvaMERG~{\scriptsize\citep{avamerg}}            & \xmark & \cmark & \cmark & \xmark & \cmark & \xmark \\
EmpathyEar~{\scriptsize\citep{empathyear}}      & \cmark & \cmark & \cmark & \xmark & \cmark & \cmark \\
\rowcolor{introgreen2!55}
\textbf{\texttt{EmpaAva}\,(Ours)}               & \cmark & \cmark & \cmark & \cmark & \cmark & \cmark \\
\bottomrule
\end{tabular}%
}
\vspace{-3mm}
\end{table}

\vspace{-2mm}
\section{Related Work}
\vspace{-2mm}
\paragraph{Empathetic Dialogue, from Text to Multimodal.}
Empathetic response generation asks a dialogue system to recognize how a user feels and reply with comfort and support. Early work lives entirely in text: EmpatheticDialogues~\citep{rashkin-et-al-2019-towards} set up a widely used benchmark, models such as MoEL, MIME, and EmpDG encode user emotion to sharpen affective quality and relevance~\citep{lin2019moel,majumder2020mime,li2020empdg}, and emotional-support dialogue adds strategy-guided comforting~\citep{esconv}. Reading and writing only words, these methods let slip the prosody, expression, speaking rate, and micro-expressions through which people reveal how they truly feel. A more recent thread therefore reaches into vision and speech~\citep{poria2019meld}, with avatar-oriented systems such as EmpathyEar and AvaMERG jointly modeling text, speech, and vision~\citep{empathyear,avamerg}. Their replies, however, still surface as text, speech, or 2D talking faces, leaving embodied feedback beyond reach.
\vspace{-2mm}
\paragraph{Talking-Face and 3D-Avatar Generation.}

Expressing empathy also calls for the right voice, facial motion, and head gesture, motivating a parallel line on avatar generation. 2D approaches such as Wav2Lip and SadTalker synthesize lip-synchronized faces from speech or a single image~\citep{wav2lip,sadtalker}, DEEPTalk injects emotionally expressive dynamics~\citep{deeptalk}, and recent diffusion-based portrait models push photorealism and head motion further~\citep{emo,vasa1,hallo}. Tied to a 2D image plane, all remain fragile on three-dimensional consistency, natural head pose, and spatial realism. 3D avatar techniques attack this directly, coupling FLAME~\citep{flame} for controllable geometry and expression with 3D Gaussian Splatting~\citep{3dgs} for high-fidelity rendering; GaussianAvatars, GaussianTalker, When Words Smile, and A$^2$-LLM showcase controllable, audio-driven 3D heads~\citep{gaussianavatars,gaussiantalker,whenwordssmile,a2llm}. The connective tissue stays missing: these systems animate a talking head from given speech, and none assembles an open, LLM-driven loop from perceiving user emotion, through empathetic planning, to embodied 3D expression. \texttt{EmpaAva} occupies exactly this space, uniting speech recognition, speech emotion recognition, LLM-based empathetic planning, structured reply plans, emotional speech synthesis, audio-driven facial motion, FLAME control, and 3DGS rendering in a single agentic system (Table~\ref{tab:related}).

\begin{figure}[!t]
\centering
\includegraphics[width=0.98\columnwidth]{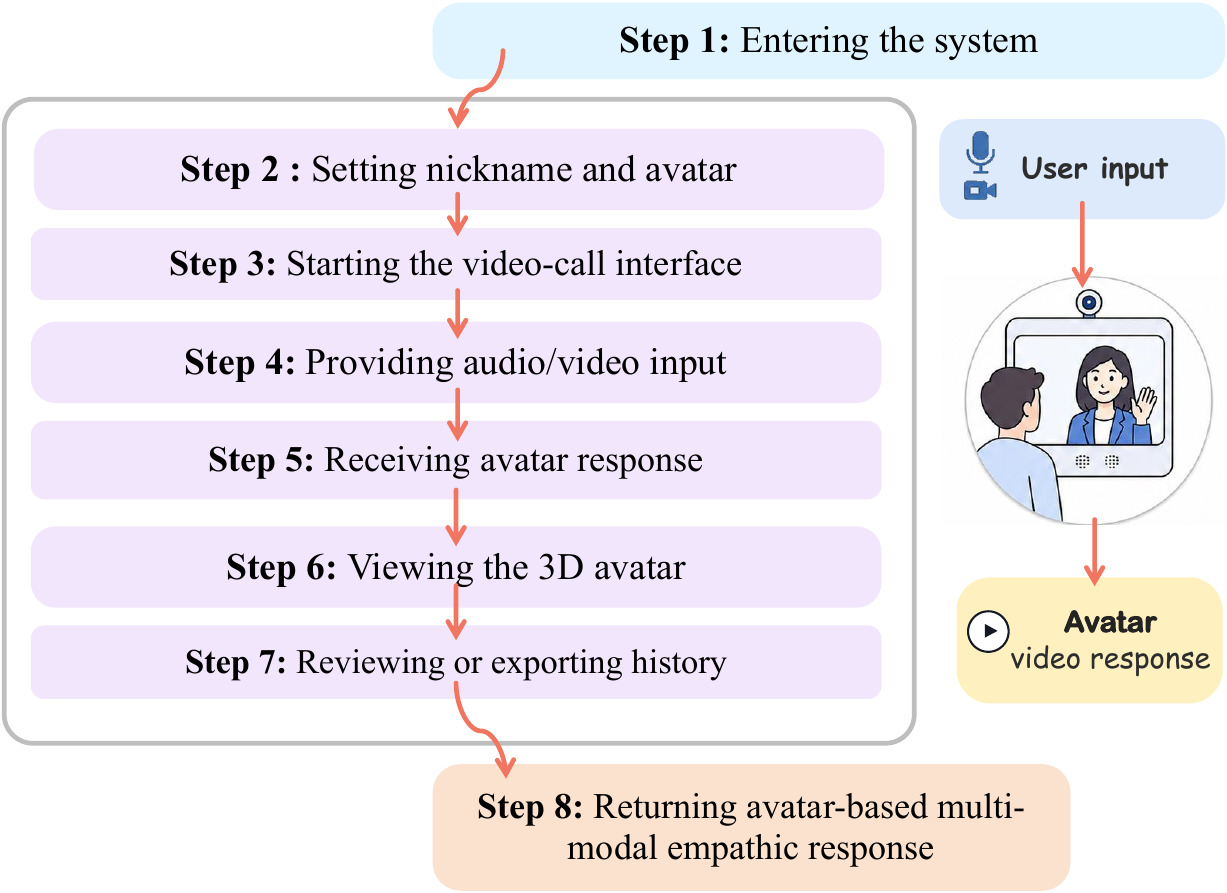}
\caption{
Workflow of the \texttt{EmpaAva} system.
}
\label{fig:workflow}
\vspace{-10pt}
\end{figure}

\begin{figure*}[!t]
\centering
\includegraphics[width=0.98\textwidth]{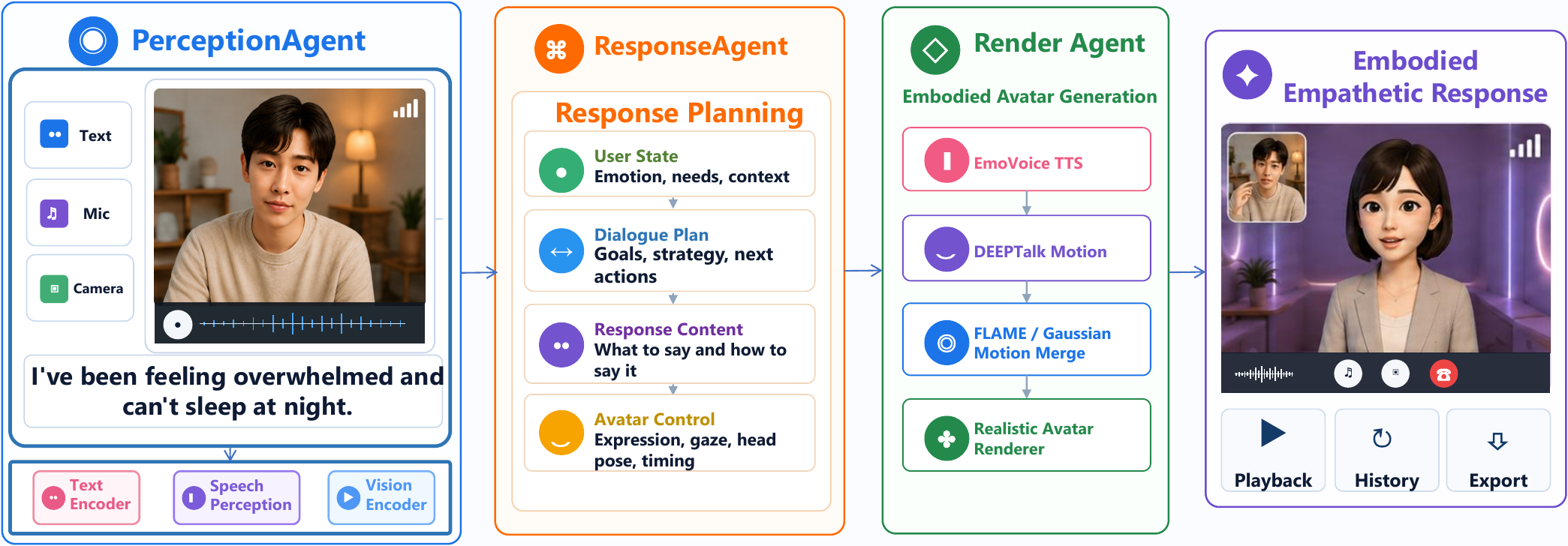}
\caption{The Tri-Agent architecture of EmpaAva. \textsc{PerceptionAgent} understands the user, \textsc{ResponseAgent} plans an empathetic reply, and \textsc{RenderAgent} turns the plan into an embodied 3D-avatar video.}
\label{fig:framework}
\vspace{-10pt}
\end{figure*}

\vspace{-2mm}
\section{System Workflow}
\label{sec:workflow}
\vspace{-2mm}
\texttt{EmpaAva} runs entirely in the browser as a video-call-like emotional booth, and Figure~\ref{fig:workflow} traces one full round of interaction, which we walk through from the user's side.

\setdefaultleftmargin{1.5em}{2em}{}{}{}{}
\begin{compactenum}

\item[$\blacktriangleright$] \textbf{Entry} (Steps 1--3).
A visitor joins as a guest with no sign-in, picks a digital avatar together with its voice and background from a pop-up, and lands straight in the call, where the avatar fills the main window and the microphone and optional camera come online.

\item[$\blacktriangleright$] \textbf{Talking} (Step 4).
The interaction mirrors a real call: microphone and camera stay on, and voice-activity detection segments each utterance and submits it automatically, so no record, send, or stop button is needed. The spoken language is detected on the fly and drives recognition, reasoning, and synthesis, letting the user converse in English or Chinese.

\item[$\blacktriangleright$] \textbf{Responding} (Steps 5--6).
For every turn, the backend reads the user's affect, plans an empathetic reply, and renders it as an embodied avatar video with synchronized emotional speech and facial motion. The reply is placed beside the user's view at matching size, and the user may replay it or rotate the 3D head to inspect the avatar from new viewpoints.

\item[$\blacktriangleright$] \textbf{History} (Steps 7--8).
The dialogue continues or stops at any moment, and each turn is archived as a paired user-and-avatar record for later replay. A lightweight registration is requested only when the user exports the full history.

\end{compactenum}

\vspace{-2mm}
\section{Implementation Specification}
\label{sec:implementation}
\vspace{-2mm}

EmpaAva turns a stream of user audio-visual input into an embodied avatar reply that can be seen, heard, and answered back. We avoid handing this whole job to a single end-to-end model, and instead factor it into three cooperating agents coordinated by an LLM (Figure~\ref{fig:framework}): \textsc{PerceptionAgent} senses the user, \textsc{ResponseAgent} decides what and how to reply, and \textsc{RenderAgent} realizes the reply as a 3D-avatar video. The agents talk to each other through a shared, human-readable state instead of opaque tensors, so every stage stays independently testable and swappable; a stronger ASR model, a larger LLM, or a higher-fidelity renderer can be dropped in without disturbing the rest, and the same transparency is what makes the pipeline easy to ablate in Section~\ref{sec:evaluation}.
\vspace{-2mm}
\subsection{PerceptionAgent}
\label{subsec:perception_module}
\vspace{-2mm}
\textsc{PerceptionAgent} converts raw user input into a structured percept. After light preprocessing (format conversion, resampling, denoising), automatic speech recognition transcribes the utterance, while speech emotion recognition reads the acoustic affect carried by prosody, pitch, and pace. That second signal matters because words alone can mislead: a flat sentence may still sound tired or anxious, and the avatar ought to answer the feeling and not only the text. When the camera is enabled, a few lightweight frames are sampled as visual context. The agent then packs the transcript, the speech-emotion label, the sampled frames, dialogue history, and input metadata into a \emph{dialogue schema} and hands it to \textsc{ResponseAgent}.
\subsection{ResponseAgent: Empathetic Response Planning}
\label{subsec:response_module}
\vspace{-2mm}
\textsc{ResponseAgent} is the decision center, and the place where our response-planning idea lives. Conditioned on the percept, the LLM first reasons over the user's emotion, its likely cause, and the conversation so far, and then produces more than an utterance. Its output is a structured \emph{reply plan} that couples the words to be spoken with how, and by whom, they should be delivered:

\begin{tcolorbox}[colback=lightyellow,colframe=introblue!65!black,boxrule=0.5pt,arc=2pt,left=5pt,right=5pt,top=3pt,bottom=3pt,fontupper=\small,title={\textsc{Reply Plan}},fonttitle=\bfseries\small,coltitle=white,colbacktitle=introblue!65!black]
\textbf{reply text}: what the avatar says~\textbullet~\textbf{emotion\,\&\,tone}: target affect and delivery~\textbullet~\textbf{avatar}: which digital human speaks~\textbullet~\textbf{voice}: speaker timbre for TTS~\textbullet~\textbf{background}: interaction scene~\textbullet~\textbf{evidence}: the user emotion and inferred cause being addressed.
\end{tcolorbox}

This plan is a single expressive contract that every downstream generator obeys, which is what keeps the comforting wording, the voice, and the facial expression aimed at the same emotional target. It also carries an empathetic strategy rather than a lone answer, typically acknowledging the feeling, naming its cause, validating it, and only then offering a low-burden suggestion. Because each field points back to the perception evidence it responds to, the plan stays grounded and interpretable, and \textsc{ResponseAgent} can revise it turn by turn as the user's state drifts.
\vspace{-2mm}
\subsection{RenderAgent}
\label{subsec:rendering_module}
\vspace{-2mm}
\textsc{RenderAgent} executes the plan and produces the avatar video. Emotional speech is first synthesized with EmotiVoice~\citep{emotivoice} under the plan's tone, so a comforting turn sounds soft and steady while an encouraging one sounds brighter. This speech then drives an audio-to-motion module that predicts frame-level FLAME~\citep{flame} parameters for jaw, lip shape, head pose, and expression. We deliberately keep appearance and motion apart: a 3D Gaussian Splatting~\citep{3dgs} head carries the identity-specific geometry and texture, FLAME acts as a low-dimensional control space, and a FLAME-to-Gaussian transfer step applies the predicted motion to the rigged Gaussian head, letting the avatar speak while holding its personalized look. Three levers govern output quality: (i) expression, jaw, and lip ranges are clamped to stay expressive without tipping into exaggeration; (ii) temporal smoothing and offset correction tighten lip-sync and suppress jitter; and (iii) 3DGS rendering with background compositing sets the avatar in a realistic call scene. The final clip carries the intended empathy across speech, motion, and appearance.

\begin{table}[!t]
\vspace{-2mm}
\centering
\fontsize{9}{10.5}\selectfont
\setlength{\tabcolsep}{1.6mm}
\caption{\raggedright
Automatic results on EmpatheticDialogues. 
Acc.: emotion-cue accuracy; Dist-1/2: diversity.
}
\resizebox{\columnwidth}{!}{
\begin{tabular}{llccc}
\hline
\multicolumn{2}{c}{\bf Models} & \bf Acc. & \bf Dist-1 & \bf Dist-2 \\
\hline

\multirow{5}{*}{\em Non-LLMs} 
& MoEL  & 3.88 & 0.67 & 2.81 \\
& MIME  & 3.52 & 0.34 & 0.96 \\
& EmpDG & 3.81 & 1.19 & 4.65 \\
& CASE  & 3.03 & 0.49 & 1.59 \\
& ESCM  & 1.58 & 0.98 & 2.38 \\
\hline

\multirow{5}{*}{\em LLMs} 
& Alpaca  & 9.38 & 3.35 & 28.05 \\
& Flan-T5 & 8.51 & 2.92 & 27.05 \\
& ChatGLM & 0.00 & 0.02 & 0.00 \\
& Qwen    & 8.73 & 3.11 & 26.10 \\
& GPT-style LLM & 6.89 & 3.30 & 24.58 \\
\hline

\multirow{3}{*}{\em Avatar-based} 
& EmpathyEar & 0.00 & 0.04 & 0.04 \\
& AvaMERG    & 8.64 & 3.30 & 28.09 \\
\cdashline{2-5}
\rowcolor{introgreen2!55} & \bf EmpaAva (Ours) 
& \bf 10.26 & \bf 4.01 & \bf 30.54 \\

\hline
\end{tabular}
}
\label{tab:text_response_comparison}
\vspace{-3mm}
\end{table}

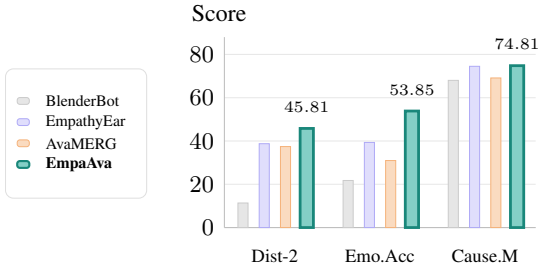
\begin{figure}[t]
\centering

\begin{minipage}[c]{0.23\linewidth}
\centering
\begin{tikzpicture}
\definecolor{bbgray}{RGB}{190,190,190}
\definecolor{eeviolet}{RGB}{166,160,245}
\definecolor{avorange}{RGB}{244,174,115}
\definecolor{oursgreen}{RGB}{105,196,185}
\definecolor{oursedge}{RGB}{25,135,122}

\node[
    draw=gray!25,
    line width=0.35pt,
    rounded corners=3pt,
    fill=white,
    inner xsep=4pt,
    inner ysep=5pt
] {
\begin{tikzpicture}
    \draw[rounded corners=1pt, fill=bbgray!38, draw=bbgray!80, line width=0.35pt]
        (0,0) rectangle (0.16,0.09);
    \node[anchor=west,font=\tiny] at (0.23,0.045) {BlenderBot};

    \draw[rounded corners=1pt, fill=eeviolet!35, draw=eeviolet!90, line width=0.35pt]
        (0,-0.28) rectangle (0.16,-0.19);
    \node[anchor=west,font=\tiny] at (0.23,-0.235) {EmpathyEar};

    \draw[rounded corners=1pt, fill=avorange!45, draw=avorange!95, line width=0.35pt]
        (0,-0.56) rectangle (0.16,-0.47);
    \node[anchor=west,font=\tiny] at (0.23,-0.515) {AvaMERG};

    \draw[rounded corners=1pt, fill=oursgreen!82, draw=oursedge, line width=0.8pt]
        (0,-0.84) rectangle (0.16,-0.75);
    \node[anchor=west,font=\tiny\bfseries] at (0.23,-0.795) {EmpaAva};
\end{tikzpicture}
};
\end{tikzpicture}
\end{minipage}
\hfill
\begin{minipage}[c]{0.74\linewidth}
\centering
\begin{tikzpicture}

\definecolor{bbgray}{RGB}{190,190,190}
\definecolor{eeviolet}{RGB}{166,160,245}
\definecolor{avorange}{RGB}{244,174,115}
\definecolor{oursgreen}{RGB}{105,196,185}
\definecolor{oursedge}{RGB}{25,135,122}

\begin{axis}[
    ybar,
    width=1.00\linewidth,
    height=4.10cm,
    ymin=0,
    ymax=88,
    symbolic x coords={Dist-2,Emo.Acc,Cause.M},
    xtick=data,
    enlarge x limits=0.24,
    ymajorgrids=true,
    grid style={draw=gray!18},
    axis line style={draw=gray!50},
    axis x line*=bottom,
    axis y line*=left,
    tick style={draw=none},
    ytick={0,20,40,60,80},
    tick label style={font=\small},
    label style={font=\small},
    xticklabel style={font=\scriptsize},
    clip=false
]

\node[
    anchor=south,
    font=\small,
    xshift=-2pt,
    yshift=2pt
] at (rel axis cs:0,1) {Score};

\addplot[
    ybar,
    bar width=3.8pt,
    bar shift=-12.0pt,
    fill=bbgray!38,
    draw=bbgray!80,
    line width=0.35pt
] coordinates {
    (Dist-2,11.35)
    (Emo.Acc,21.75)
    (Cause.M,68.01)
};

\addplot[
    ybar,
    bar width=3.8pt,
    bar shift=-4.0pt,
    fill=eeviolet!35,
    draw=eeviolet!90,
    line width=0.35pt
] coordinates {
    (Dist-2,38.75)
    (Emo.Acc,39.33)
    (Cause.M,74.51)
};

\addplot[
    ybar,
    bar width=3.8pt,
    bar shift=4.0pt,
    fill=avorange!45,
    draw=avorange!95,
    line width=0.35pt
] coordinates {
    (Dist-2,37.43)
    (Emo.Acc,30.97)
    (Cause.M,69.13)
};

\addplot[
    ybar,
    bar width=5.0pt,
    bar shift=12.0pt,
    fill=oursgreen!82,
    draw=oursedge,
    line width=1.05pt,
    nodes near coords={
        \pgfmathprintnumber[
            fixed,
            fixed zerofill,
            precision=2
        ]{\pgfplotspointmeta}
    },
    point meta=y,
    every node near coord/.append style={
        font=\tiny\bfseries,
        anchor=south,
        text=black,
        yshift=2pt
    }
] coordinates {
    (Dist-2,45.81)
    (Emo.Acc,53.85)
    (Cause.M,74.81)
};

\end{axis}
\end{tikzpicture}
\end{minipage}
\vspace{-3mm}
\caption{Automatic evaluation on the AvaMERG test set. Higher scores indicate better response diversity, emotion understanding, and emotion-cause modeling.}
\label{fig:system_comparison_auto}
\vspace{-3mm}
\end{figure}

\vspace{-2mm}
\section{Evaluation and Analysis}
\label{sec:evaluation}
\vspace{-2mm}
We evaluate \texttt{EmpaAva} on text-level response generation, end-to-end multimodal interaction, and embodied expression quality, plus two multi-turn case studies. Subjective scores are averaged over human raters on a five-point scale.

\begin{table}[!t]
\centering
\small
\setlength{\tabcolsep}{3.5pt}
\caption{Human evaluation scores from a questionnaire study with 10 participants, each evaluating 5 user turns with four anonymized system responses. Emp.: Response Empathy; Rel.: Relevance; Spec.: Specificity; Pref.: Overall Preference votes out of 50.}
\label{tab:system_comparison_human}
\begin{tabular}{lcccc}
\toprule
\textbf{Method} & \textbf{Emp.} & \textbf{Rel.} & \textbf{Spec.} & \textbf{Pref.} \\
\midrule
BlenderBot      & 4.36 & 4.53 & 4.17 & 7 \\
EmpathyEar      & 4.36 & 4.47 & 4.25 & 12 \\
AvaMERG         & 4.18 & 4.43 & 4.07 & 13 \\
\rowcolor{introgreen2!55}
\textbf{EmpaAva (Ours)}  & \textbf{4.42} & \textbf{4.65} & \textbf{4.35} & \textbf{18} \\
\bottomrule
\end{tabular}
\end{table}

\begin{table}[!t]
\centering
\small
\setlength{\tabcolsep}{4.8pt}
\caption{Automatic evaluation of avatar embodied expression quality on the AvaMERG evaluation set. Higher LSE-C and A-V Cos are better, while lower LSE-D is better. Best results are in bold. The underlined value denotes the second-best LSE-D score.}
\label{tab:avatar_quality_real_metrics}
\begin{tabular}{lrrr}
\toprule
\textbf{Method} & \textbf{LSE-C}$\uparrow$ & \textbf{LSE-D}$\downarrow$ & \textbf{A-V Cos}$\uparrow$ \\
\midrule
TTS-only                       & 0.416 & 16.846 & 0.274 \\
Wav2Lip               & 6.452 & \textbf{7.940} & 0.339 \\
SadTalker             & 3.727 & 11.186 & 0.226 \\
DEEPTalk             & 1.432 & 13.491 & 0.338 \\
3DGS                 & 0.826 & 16.526 & 0.294 \\
\rowcolor{introgreen2!55}
EmpaAva (Ours)        & \textbf{7.649} & \underline{8.242} & \textbf{0.346} \\
\bottomrule
\end{tabular}
\end{table}

\vspace{-2mm}
\subsection{Empathetic System Comparison}
\label{subsec:system_comparison}
We compare \texttt{EmpaAva} at the text and end-to-end embodied-avatar levels.

\vspace{-2mm}
\subsubsection{Text-Level Response Comparison}
\label{subsubsec:text_response_comparison}
\paragraph{Protocol.}
We first evaluate EmpaAva's text response module on the full EmpatheticDialogues test split~\citep{rashkin-et-al-2019-towards} (5,255 samples with emotion labels and references), one response per system per sample. Baselines cover non-LLM ERG methods (MoEL, MIME, EmpDG, CASE, ESCM), LLMs (Alpaca, Flan-T5, ChatGLM, Qwen, and a GPT-style LLM), and text modules of multimodal systems (EmpathyEar, AvaMERG). For EmpaAva, we use the same deterministic decoding setting for all test samples.

\paragraph{Results.}
Table~\ref{tab:text_response_comparison} shows EmpaAva leading on all three metrics (Acc.\ 10.26, Dist-1 4.01, Dist-2 30.54), indicating stronger emotion-cue alignment and lexical diversity than non-LLM, LLM, and avatar-based baselines.

\begin{figure*}[!t]
\centering
\includegraphics[width=0.95\textwidth]{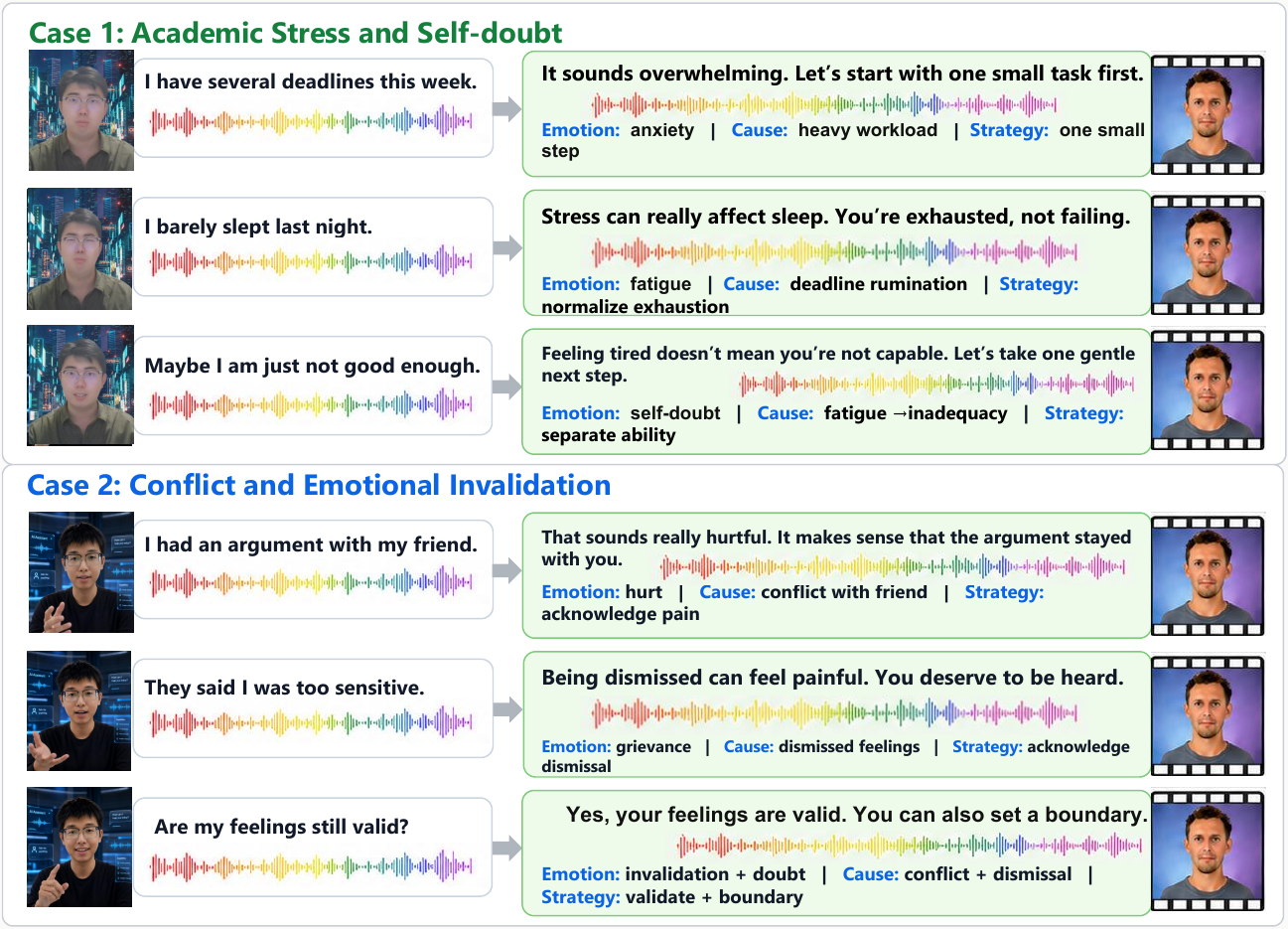}
\vspace{-2mm}
\caption{Qualitative multi-turn case studies of \texttt{EmpaAva}.}
\label{fig:case_study}
\vspace{-3mm}
\end{figure*}

\subsubsection{End-to-End Avatar Comparison}
\label{subsubsec:end_to_end_system_comparison}
\paragraph{Protocol.}
We evaluate EmpaAva's complete multimodal output (text, emotional speech, 3D avatar video) on the AvaMERG test set~\citep{avamerg}, against BlenderBot~\citep{blenderbot} with neutral TTS and avatar, EmpathyEar~\citep{empathyear} with 2D talking-face, and AvaMERG~\citep{avamerg}.
We report automatic metrics for diversity, emotion understanding, and cause modeling, plus a questionnaire where 10 participants each rate 5 turns over four anonymized responses (50 preference votes); MECS and AV-Agree serve as cross-modal diagnostics (Appendix~\ref{app:e2e_eval_details}).

\vspace{-2mm}
\paragraph{Results.}
In Figure~\ref{fig:system_comparison_auto}, EmpaAva attains the best Dist-2 (45.81), Emo.Acc (53.85), and Cause.M (74.81); on 100 complete avatar outputs it further reaches 22.0\% MECS and 40.0\% AV-Agree. Human evaluation in Table~\ref{tab:system_comparison_human} echoes the trend: highest Empathy (4.42), Relevance (4.65), Specificity (4.35), and preference (18/50).


\subsection{Avatar Embodied Expression Quality}
\label{subsec:avatar_quality}
\paragraph{Protocol.}
This experiment isolates avatar generation, fixing response text, target emotion, speech input, avatar identity, and video format. On 2,308 AvaMERG samples, we compare TTS-only playback, Wav2Lip~\citep{wav2lip}, SadTalker~\citep{sadtalker}, DEEPTalk~\citep{deeptalk}, a 3DGS Avatar~\citep{gaussianavatars}, and EmpaAva, reporting SyncNet LSE-C/LSE-D for lip synchronization and A-V Cos for speech-expression emotional consistency.

\vspace{-2mm}
\paragraph{Results.}
In Table~\ref{tab:avatar_quality_real_metrics}, EmpaAva obtains the highest LSE-C (7.649) and A-V Cos (0.346). Wav2Lip wins LSE-D (7.940) via explicit lip-sync optimization, yet EmpaAva stays close (8.242) with better emotional consistency, giving the best overall sync--affect balance.

\subsection{Qualitative Multi-turn Case Study}
\label{subsec:case_study}

\paragraph{Case 1: Academic Stress and Self-doubt.}
\vspace{-1mm}
The user (Figure~\ref{fig:case_study}, top) gradually reveals assignment pressure, poor sleep, and self-doubt; EmpaAva tracks this drift, links it to the earlier stressors, and separates exhaustion from ability.

\paragraph{Case 2: Interpersonal Conflict and Emotional Invalidation.}
\vspace{-1mm}
The user (Figure~\ref{fig:case_study}, bottom) is hurt and called ``too sensitive''; EmpaAva names the invalidation, validates the feelings, reduces self-blame, and suggests boundary communication without judging the friend.

\section{Conclusion}
\vspace{-2mm}
We presented EmpaAva, an open-source agentic 3D-avatar chatbot for live, face-to-face empathetic dialogue: its LLM-driven \textbf{Tri-Agent Architecture} closes the loop from affect perception to embodied rendering, and \textbf{Response Planning} keeps speech, facial motion, and rendering on one empathetic intent. Released with an online demo, it offers an inspectable testbed for embodied empathetic agents; richer bodily expression, longer emotional memory, and lower latency remain future work.

\newpage

\section*{Ethical Considerations and Broader Impact}
\label{app:ethics}

\paragraph{Privacy, consent, and data handling.}
EmpaAva takes microphone and optional camera input through a video-call-like interface, and both channels are strictly opt-in: the browser requests each permission explicitly, the purpose of every permission is stated before interaction, and the system remains fully functional with the camera disabled. Audio is used only for speech recognition and speech emotion recognition, and sampled video frames only for turn-level multimodal perception; neither stream is used to train models or shared with any third party. The research demo may produce intermediate artifacts, including perception results, reply plans, synthesized speech, facial-motion parameters, and rendered videos, solely for debugging, reproducibility, history review, and user-initiated export. All examples shown in this paper are selected and anonymized with the consent of the recorded participants, and expose no real user identities, raw recordings, or sensitive information.

\paragraph{Embodied presence and identity safety.}
A photorealistic talking avatar creates stronger social presence than a chat window, which raises risks of deception and emotional over-reliance. EmpaAva therefore always presents itself as an AI companion and never impersonates a real person. Avatar identities and TTS voices derive from publicly released research assets and synthetic sources; the system provides no capability for cloning the face or voice of a user or any third party. We further avoid manipulative emotional cues and engagement-maximizing behaviors, and the interface allows the user to stop or leave a session at any moment.

\paragraph{Not a substitute for professional care.}
EmpaAva is a companion-oriented research demonstration, not a medical or psychological service; its prompts forbid clinical or diagnostic advice and steer replies toward warm, validating, low-burden support. When a conversation signals severe distress or self-harm, the system avoids prescriptive language and encourages the user to seek trusted people or professional help. Any deployment in real wellbeing contexts would require clinical oversight and safety review beyond this demo.

\paragraph{Human evaluation.}
The questionnaire study involved voluntary on-campus participants who were informed of the research purpose, rated fully anonymized system outputs presented in randomized order, and contributed no personal data beyond their ratings.

\paragraph{Responsible release.}
We release the code and demo for research purposes under terms that prohibit deceptive impersonation, harassment, and unauthorized use of anyone's likeness or voice. We also document known limitations, e.g., speech emotion recognition can misread affect and generated comfort can be generic, so that downstream users can assess risks before adoption.

\bibliography{anthology}

\newpage

\appendix

\section{System Implementation Details}
\label{app:impl}

Table~\ref{tab:app_modules} lists the backbone of each Tri-Agent module. EmpaAva runs in the browser: a lightweight frontend renders the call booth and streams audio/video input, while the backend orchestrates the three agents and returns the rendered clip.

\begin{table}[h]
\centering
\scriptsize
\setlength{\tabcolsep}{3pt}
\renewcommand{\arraystretch}{1.15}
\caption{Backbone modules of EmpaAva.}
\label{tab:app_modules}
\begin{tabular}{
@{}p{0.22\columnwidth}
p{0.34\columnwidth}
p{0.36\columnwidth}@{}
}
\toprule
\textbf{Stage} & \textbf{Module} & \textbf{Backbone} \\
\midrule
\rowcolor{gray!12}
Perception & Speech recognition (ASR) & Whisper-small \\
Perception & Speech emotion recognition (SER) & emotion2vec / FunASR \\
\rowcolor{gray!12}
Response & Reasoning \& planning LLM & OpenAI-compatible LLM / AvaMERG fallback \\
Render & Emotional TTS & EmotiVoice~\citep{emotivoice} \\
\rowcolor{gray!12}
Render & Audio-to-motion & DEEPTalk~\citep{deeptalk} \\
Render & Face control space & FLAME~\citep{flame} \\
\rowcolor{gray!12}
Render & Avatar rendering & 3DGS / GaussianAvatars~\citep{3dgs,gaussianavatars} \\
\bottomrule
\end{tabular}
\vspace{-2mm}
\end{table}

\paragraph{Deployment and latency.}
The backend uses FastAPI with persistent local workers. On an NVIDIA H200, warm-start per-turn latency averages 45.8s, dominated by Gaussian rendering/export (41.5s), with 2.1s perception, 0.002s planning, 0.27s TTS, and 0.05s audio-to-motion.


\section{Agent Prompt Designs}
\label{app:prompts}

We expose the full agent prompt templates; angle-bracket placeholders (e.g., \texttt{<user-text>}) are filled at run time from the dialogue schema.

\paragraph{Global system role.}
\begin{promptbox}{System Prompt (shared)}
You are the reasoning core of EmpaAva, an empathetic 3D-avatar
companion. Your goal is to understand how the user feels, why
they feel that way, and to respond with warmth, validation, and
gentle, low-burden support. Never give clinical or diagnostic
advice. Always keep the reply short enough to be spoken aloud
by an avatar in a video call.
\end{promptbox}

\paragraph{PerceptionAgent: state fusion.}
This prompt fuses transcript, acoustic emotion, and optional visual cues into a compact user-state summary.
\begin{promptbox}{PerceptionAgent Prompt}
Inputs:
- transcript: <user-text>
- speech emotion (SER): <speech-emotion>
- visual cues (optional): <face-cues>
- dialogue history: <history>

Summarize the user's current state in 1-2 sentences:
(a) the dominant emotion, (b) its most likely cause, and
(c) whether it intensifies or eases relative to prior turns.
Return only the summary.
\end{promptbox}

\paragraph{ResponseAgent: empathetic planning.}
It reasons over the fused state and emits a structured \emph{reply plan} (Appendix~\ref{app:schema}) rather than a bare utterance.
\begin{promptbox}{ResponseAgent Prompt}
User state: <perception-summary>
Speech emotion: <speech-emotion>
Dialogue history: <history>
Available avatars: <avatar-list>
Available voices: <voice-list>
Available backgrounds: <bg-list>

Think step by step, then produce a reply plan:
1. Name the user's emotion and its likely cause.
2. Choose an empathetic strategy: acknowledge -> address
   cause -> validate -> offer one low-burden suggestion.
3. Write the reply text to be spoken (up to 3 sentences).
4. Set the target emotion and tone, avatar, voice, and
   background for the delivery.
5. Cite the perception evidence each choice responds to.

Output the reply plan as a single JSON object following
the schema. Do not add any text outside the JSON.
\end{promptbox}


\section{Reply Plan Schema}
\label{app:schema}

The reply plan is the machine-readable contract passed from \textsc{ResponseAgent} to \textsc{RenderAgent}, specifying the spoken response, delivery style, selected avatar, TTS speaker, background, and links to intermediate perception artifacts.

\paragraph{Example reply plan.}

{\footnotesize
\begin{promptbox}{Reply Plan Example}
\raggedright
\{"schema": 1,\\
"agent": "ResponseAgent",\\
"strategy": "acknowledge\_and\_support",\\
"emotion": "anxiety",\\
"cause": "multiple approaching deadlines",\\
"evidence": "I have several deadlines this week.",\\
"reply\_text": "That sounds overwhelming. Let's take one small step first.",\\
"tone": "warm",\\
"expression": "concerned",\\
"selected\_avatar\_id": "306",\\
"selected\_tts\_speaker\_id": "6224",\\
"background": null\}\\
\end{promptbox}
}


\section{Avatar Identity Gallery}
\label{app:viz}

Figure~\ref{fig:app_avatar_gallery} shows representative renderings by EmpaAva, demonstrating visually consistent 3D digital humans with diverse identities and expressions.

\begin{center}
\includegraphics[width=0.99\columnwidth]{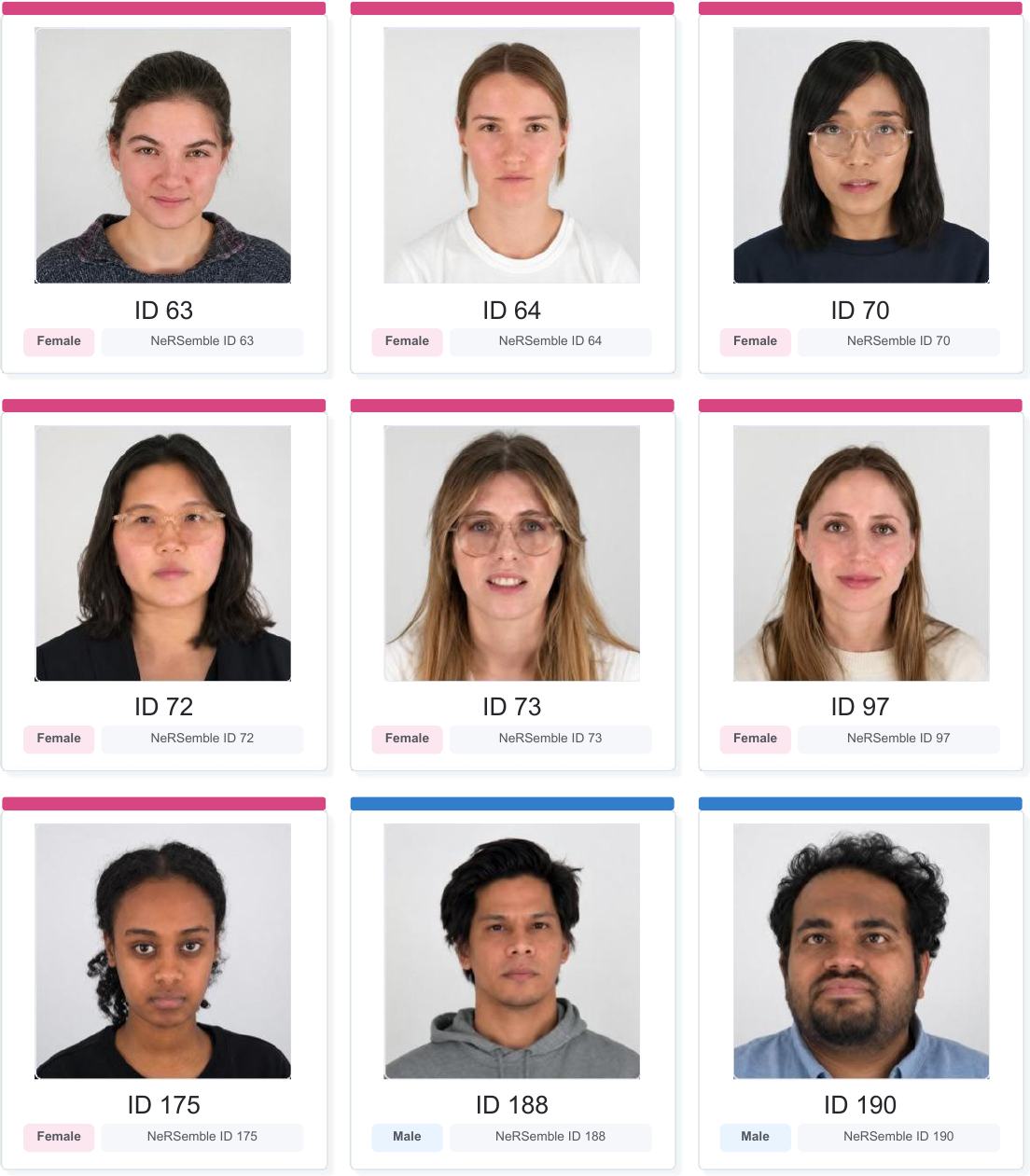}
\captionof{figure}{Qualitative avatar renderings across different identities.}
\label{fig:app_avatar_gallery}
\end{center}

\section{Experimental Settings}
\label{app:settings}

\paragraph{Datasets.}
We use AvaMERG and EmpatheticDialogues as complementary evaluation sources. 
AvaMERG provides 6,288 valid multimodal instances after filtering, covering user input, emotions, responses, and avatar resources, and serves both response-level and avatar-level evaluation. 
EmpatheticDialogues contains about 25K text conversations with 32 emotion labels and is used for text-response and response-planning evaluation. 
We sample from both datasets with balanced emotion distribution, remove invalid cases, and build the avatar evaluation subset from AvaMERG by selecting samples with valid text, emotion labels, avatar identities, and renderable resources.

\paragraph{Human evaluation protocol.}
We conduct a small pilot human evaluation with on-campus students from NLP, multimodal learning, or psychology backgrounds. 
Participants view anonymized system responses in randomized order with matched player size and normalized audio, rate \emph{Response Empathy}, \emph{Relevance}, and \emph{Specificity} on a 1--5 Likert scale, and select one overall preferred response. 
A representative rater instruction follows.

\begin{promptbox}{Rater Instruction (excerpt)}
You will watch a short clip of a user turn and the avatar's
response. Rate 1--5 on each dimension:

- Response Empathy: is the reply warm, validating, and
  supportive?

- Relevance: does the reply directly address the user's
  situation and emotional state?

- Specificity: does the reply provide concrete and
  context-specific support rather than a generic response?

After rating the four anonymized system responses for the
same user turn, select the one you prefer overall.

Judge only the presented response; ignore minor differences
in video resolution or audio volume.
\end{promptbox}


\paragraph{Evaluation metrics.}
\begingroup
\footnotesize
\setlength{\abovedisplayskip}{2pt}
\setlength{\belowdisplayskip}{2pt}
\setlength{\jot}{2pt}

\begin{align}
\mathrm{Acc}
&= 100 \times \frac{1}{N}\sum_{i=1}^{N}
\mathbb{I}\bigl[\hat{y}_i^{e}=y_i^{e}\bigr], \\
\mathrm{Dist}\text{-}n
&= 100 \times
\frac{
\bigl|\bigcup_i \operatorname{ngram}_n(r_i)\bigr|
}{
\sum_i \bigl|\operatorname{ngram}_n(r_i)\bigr|
},
\quad n\in\{1,2\}, \\
\mathrm{Cause.M}
&= 100 \times \frac{1}{N}\sum_i
\mathbb{I}\bigl[\hat{y}^{c}_i = y^{c}_i\bigr], \\
\mathrm{MECS}
&= 100 \times \frac{1}{N}\sum_i
\mathbb{I}\bigl[
\hat{y}^{t}_i = \hat{y}^{a}_i
= \hat{y}^{v}_i = y_i
\bigr], \\
\mathrm{Score}(m)
&= \frac{1}{JN}\sum_{j=1}^{J}\sum_{i=1}^{N}s_{j,i}^{m}, \\
\mathrm{LSE}\text{-}\mathrm{D}
&= \tfrac{1}{M}
\sum_{i,t}
\|a_{i,t}-v_{i,t}\|_2, \\
\mathrm{LSE}\text{-}\mathrm{C}
&= \tfrac{1}{M}
\sum_{i,t}
\bigl(
\operatorname{med}_{\tau\ne 0}
\|a_{i,t}-v_{i,t+\tau}\|_2 \nonumber\\
&\qquad
-\|a_{i,t}-v_{i,t}\|_2
\bigr).
\end{align}

\endgroup

\paragraph{Full automatic results.}
\label{app:e2e_eval_details}
Table~\ref{tab:app_system_comparison_auto} reports the complete numerical automatic results corresponding to Figure~\ref{fig:system_comparison_auto}.

\begin{table}[!h]
\centering
\small
\setlength{\tabcolsep}{4.5pt}
\caption{Complete numerical results for end-to-end automatic evaluation on the \texttt{AvaMERG} test set. Dist-2: response diversity; Emo.Acc: emotion understanding accuracy; Cause.M: emotion-cause modeling.}
\label{tab:app_system_comparison_auto}
\begin{tabular}{lccc}
\toprule
\textbf{Method} & \textbf{Dist-2↑} & \textbf{Emo.Acc↑} & \textbf{Cause.M↑} \\
\midrule
BlenderBot     & 11.35 & 21.75 & 68.01 \\
EmpathyEar     & 38.75 & 39.33 & 74.51 \\
AvaMERG         & 37.43 & 30.97 & 69.13 \\
\rowcolor{introgreen2!55}
EmpaAva (Ours)  & \textbf{45.81} & \textbf{53.85} & \textbf{74.81} \\
\bottomrule
\end{tabular}
\end{table}

\end{document}